\documentclass[10pt,twocolumn,letterpaper]{article}

\usepackage{cvpr}              

\usepackage{graphicx}
\usepackage{amsmath}
\usepackage{amssymb}
\usepackage{booktabs}
\usepackage[accsupp]{axessibility} 
\usepackage[pagebackref,breaklinks,colorlinks]{hyperref}

\usepackage[capitalize]{cleveref}
\crefname{section}{Sec.}{Secs.}
\Crefname{section}{Section}{Sections}
\Crefname{table}{Table}{Tables}
\crefname{table}{Tab.}{Tabs.}

\def\cvprPaperID{*****} 
\def\confName{CVPR}
\def\confYear{2022}

\begin{document}

\title{Transformer-based Multimodal Information Fusion for Facial Expression Analysis}
\author{Wei Zhang, 
Feng Qiu, 
Suzhen Wang, 
Hao Zeng, 
Zhimeng Zhang, \\
Rudong An,
Bowen Ma, 
Yu Ding\footnotemark[2]\thanks{Corresponding Author.}\\
Virtual Human Group, Netease Fuxi AI Lab\\
{\tt\small \{zhangwei05,qiufeng,wangsuzhen,zenghao03,zhangzhimeng,}\\
{\tt\small anrudong,mabowen01,dingyu01\}@corp.netease.com}\\
}
\maketitle


\begin{abstract}

Human affective behavior analysis has received much attention in human-computer interaction (HCI). In this paper, we introduce our submission to the CVPR 2022 Competition on Affective Behavior Analysis in-the-wild (ABAW). 
To fully exploit affective knowledge from multiple views, we utilize the multimodal features of spoken words, speech prosody, and facial expression, which are extracted from the video clips in the \textit{Aff-Wild2} dataset. Based on these features, we propose a unified transformer-based multimodal framework for Action Unit detection and also expression recognition. Specifically, the static vision feature is first encoded from the current frame image. At the same time, we clip its adjacent frames by a sliding window and extract three kinds of multimodal features from the sequence of images, audio, and text. 
Then, we introduce a transformer-based fusion module that integrates the static vision features and the dynamic multimodal features. 
The cross-attention module in the fusion module makes the output integrated features focus on the crucial parts that facilitate the downstream detection tasks.
We also leverage some data balancing techniques, data augmentation techniques, and postprocessing methods to further improve the model performance.
In the official test of ABAW3 Competition, our model ranks first in the EXPR and AU tracks. The extensive quantitative evaluations, as well as ablation studies on the \textit{Aff-Wild2}  dataset, prove the effectiveness of our proposed method.

\end{abstract}

\section{Introduction}
\label{sec:intro}
Human affective behavior analysis plays a significant role in human-computer interaction. The basic expressions, Action Units (AU), and Valence-Arousal (VA), as three commonly-used expression representations, are usually used to infer human emotions. AU explains human expression from the view of facial muscle movement based on the Facial Action Coding System (FACS)~\cite{ekman}. The basic expression uses discrete and explicit definitions to represent human expressions. Valence and Arousal are two continuous values used to describe human emotion states. 
However, most existing datasets ~\cite{jaffe, zhang2014bp4d,zhang2016multimodal,mollahosseini2017affectnet} contain only one of the three representations. 
Different from them, \textit{Aff-Wild}~\cite{zafeiriou2017aff,kollias2020analysing,kollias2019deep,kollias2019face} and \textit{Aff-Wild2}~\cite{kollias2019expression,kollias2021affect,kollias2021distribution,kollias2021analysing,kollias2022abaw} containing all the three representation labels have received considerable attention from both academic and industrial communities. 
Also, the videos in the \textit{Aff-Wild2}~\cite{kollias2019expression,kollias2021affect,kollias2021distribution,kollias2021analysing,kollias2022abaw} show human spontaneous affective behaviors in the wild, pushing the affective analysis to fit with the real-world scenarios.

\begin{figure}
    \centering
    \includegraphics[ width=1\linewidth]{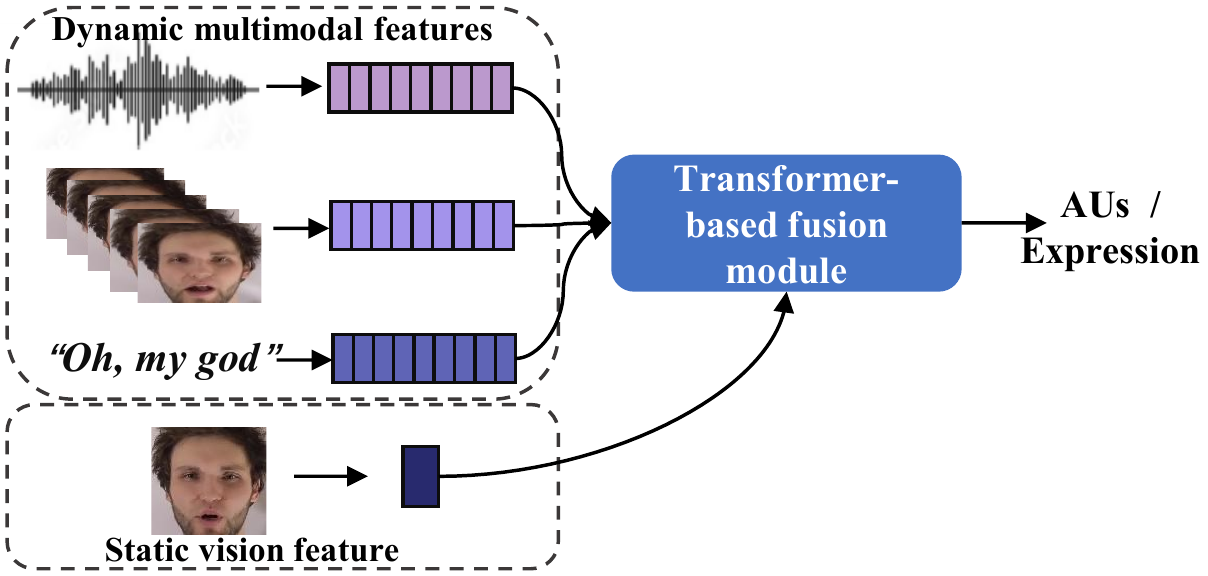}
    \caption{The sketch of our proposed transformer-based multimodal information fusion framework for AU detection and expression recognition. The transformer-based fusion module is used to integrate the static vision feature and dynamic multimodal features.}
    \label{fig:intro}
\end{figure}

Considering the fact that visual expression, spoken words, and speech prosody also imply rich emotion information, we propose a transformer-based multimodal information fusion framework for AU detection and expression recognition~(See~\cref{fig:intro}). 
First, a static vision feature extractor is used to capture the vision expression feature of the static image. 
Then we employ the sliding windows to obtain the sequence of adjacent frames, audio, and text. Based on the three kinds of multimodal signals, we capture the corresponding prior features by some pretrained models or open-sourced tools.
In particular, we employ the pretrained DLN~\cite{zhang2021learning} to extract the expression embedding feature of the cropped faces sequence. The sequence of audio features, extracted by the open-sourced librosa tool~\cite{mcfee2015librosa}, consists of MFCC, pitch, and short-time energy. Also, we obtain the spoken words by ASR tools~\cite{speechbrain} and produce their word embedding based on the pretrained Bert~\cite{devlin2018bert}. 

After obtaining these features, the dynamic uni-modal features combined with the static vision feature are fed into a transformed-based fusion module.
The static vision feature is injected into the multi-head cross attention module as $K$ and $V$ for attention coefficient computation.
In this way, we can effectively integrate the static vision features and dynamic multimodal features. The integrated multimodal features are forced to pay more attention to key point parts for the current frame detection task.
In addition, to alleviate the serious imbalance in the expression recognition 
trainset, we resample the training samples based on the number of categories and video frames. Remix~\cite{chou2020remix} data augmentation is also used to relieve the imbalance. Considering that the expression labels seldom change rapidly in continuous sequence frames, we propose the filter strategy to smooth the final prediction results. 


In sum, the contributions of this work are three-fold:
\begin{itemize}
    \item To fully exploit the in-the-wild emotion information, we utilize the multimodal information from the images, audio and text and propose a unified multimodal framework for AU detection and expression recognition. With the valid prior multimodal information, our model realizes the effective affective analysis from different views.
    \item We introduce a transformer-based fusion module for integrating the static vision feature and dynamic multimodal features. With the multi-head cross-attention, the output integrated features capture the crucial information for affective analysis. In the ABAW3 competition, our method ranks first in both AU and expression tracks. The final test set score and quantitative experiments can prove the superiority of our method.  
\end{itemize}







\section{Related works}
In this section, we will introduce some related works about the facial expression representations, AU detection, and expression recognition.

\subsection{Facial expression representations}
The main challenges in the 3rd Workshop and Competition on Affective Behavior Analysis in-the-wild (ABAW) are about the commonly-used facial expression representations: AU, basic expression categories, and Valence-Arousal. 
Ekman \textit{et al.}~\cite{ekman} first introduces the definition of the AUs from the Facial Action Coding System (FACS), which contains 32 atomic facial action descriptors based on the movements of the facial muscles.
The basic expression categories describe human affection by seven discrete semantic definitions, namely Anger, Disgust, Fear, Happiness, Sadness, Surprise, and Neutral.
Valence-Arousal describes the emotion state of humans by two continuous values. Valence indicates how positive or negative is the human affective behavior and Arousal indicates how active or passive is the human affective behavior. The values of Valence and Arousal are between -1 and 1.
Except for the above representations, recently a compact and continuous expression embedding~\cite{vemulapalli2019compact} is proposed to represent fine-grained human expressions. The main idea of it is to reserve the expression similarity on a low dimensional space.


\subsection{AU detection}
In the area of AU detection, one of the main challenges is the bad generalization caused by the limited identities in the commonly-used AU datasets.
Therefore, numerous works focus on introducing more additional information to enhance the detection task. To facilitate the model with local features, multi-task methods usually combine AU detection with landmark detection~\cite{benitez2017recognition, shao2021jaa} or landmark-based attention map prediction~\cite{jacob2021facial}.  SEV-Net~\cite{yang2021exploiting} proposes to utilize the textual descriptions of local details to generate a regional attention map.

In ABAW2 competition, multi-task~\cite{zhang2021prior,jin2021multi,thinh2021emotion,vu2021multitask} that combines the tasks of expression recognition or VA detection is often employed to add underlying features for AU task. Zhang \textit{et al.}~\cite{zhang2021prior} pretrain the backbone on the DLN~\cite{zhang2021learning} and propose a streaming multi-task framework, winning first place in AU and EXPR tracks. Jin \textit{et al.}~\cite{jin2021multi} with multi-task framework also introduce the multimodal information from audio and visual signals, winning second place in AU and EXPR tracks.

In ABAW3 competition, Jiang \textit{et al.}~\cite{AU_2} pretrain the feature extractor on Glint360K and some private commercial datasets before conducting the AU detection task in the ABAW3. They achieve second place in the AU track. 
Savchenko \textit{et al.}~\cite{EXPR4} also pretrain the backbone on large facial expression datasets and finetune it for AU detection or expression recognition. They rank 5-th in the AU track and 4-th in the EXPR track. Transformer~\cite{vaswani2017attention} is also often used. Wang \textit{et al.}~\cite{AU_4} proposed a two-stream transformer-based framework to model the inherent relationships of AUs. Tallec \textit{et al.}~\cite{AU_6} and Nguyen \textit{et al.}~\cite{AU_3} use the transformer-based architecture for the multi-label AU detection, taking advantage of the self-attention and cross-attention modules to learn the local features for each action unit. Multimodal information like audio is also introduced in AU task~\cite{AU_4}.



\begin{figure*}
    \centering
    \includegraphics[ width=1\linewidth]{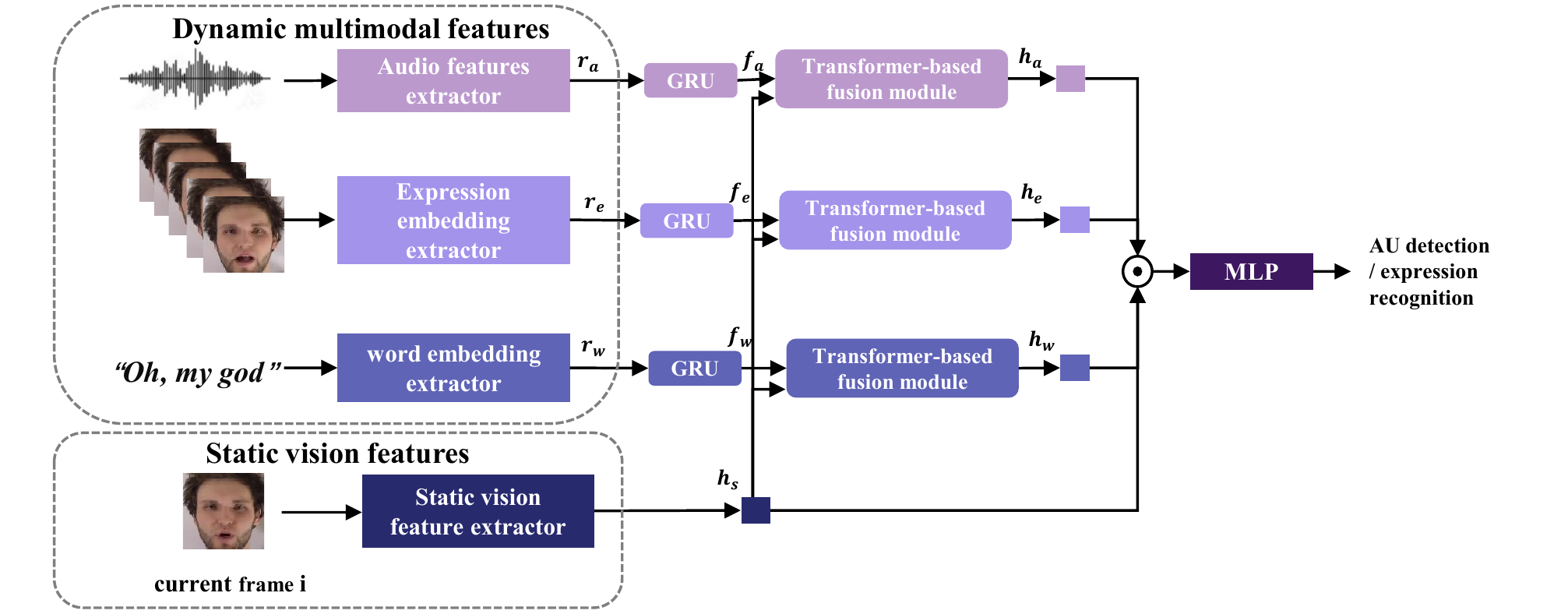}
    \caption{The pipeline of our proposed multimodal framework for AU detection and expression recognition. The static vision feature extractor captures the static vision expression features $h_s$ from the current frame. Three dynamic multimodal feature extractors with the multimodal sequence information as input focus on providing emotion features from different views. The transformer-based fusion module is used to incorporate the static and dynamic multimodal features. The output features from three transformer-based fusion modules concatenated with $h_s$ are sent into the multilayer perceptron~(MLP) for detection tasks. }
    \label{fig:pipeline}
\end{figure*}

\subsection{Expression recognition}
Expression recognition aims to identify the basic expression categories of humans, such as anger, disgust, fear, happiness, sadness, surprise, etc.
Recently, benefiting from the development of the neural network, deep-learning-based approaches are applied to end-to-end expression recognition. 
Some works \cite{breuer2017deep, jung2015joint} employ CNN to extract the spatial and temporal features and improve expression recognition. Some other works \cite{kim2017multi, chu2017learning} utilize LSTM to model the temporal dependencies and characteristics within consecutive facial images. To focus on the most relevant features, some recent works introduce the attention mechanism into CNN \cite{cheng2021multimodal, Multistage, MARN, MFN}.

In the ABAW2 Competition, the winner in EXPR track \cite{zhang2021prior} leverages an identity-invaraint expression feature and develops a multi-task streaming network to extract interaction relationships among three modal representations. Thinh \textit{et al.}~\cite{thinh2021emotion} takes advantage of the multi-task learning technique by combining the knowledge for two correlated tasks, AUs prediction and emotion classification. Deng \textit{et al.}~\cite{deng2021iterative} is dedicated to solving the problem of emotion uncertainty and proposes a framework that predicts both emotions labels and the estimated uncertainty. Then, this work improves emotion recognition and uncertainty estimation by the self-distillation algorithm. Jin \textit{et al.}~\cite{jin2021multi} use both AU and expression annotations to train the model and apply a sequence model to further extract associations between frames in sequences. Wang \textit{et al.}~\cite{wang2021multi} aims to tackle the problem of incomplete labeled datasets in multi-task affective behavior recognition methods. This approach trains a semi-supervised model as the teacher network to predict pseudo labels for unlabeled data. Then these pseudo labels are utilized for student network training, which allows it to learn from unlabeled data.

In the ABAW3 Competition, other teams also propose effective frameworks. 
Jeong \textit{et al.}~\cite{EXPR2} use the affinity loss to train the feature extractor from the images. And a multi-head attention network in an orchestrated fashion is proposed to extract diverse attention for expression recognition.
To distinguish the similar expressions, Xue \textit{et al.}~\cite{EXPR3} propose a two-stage framework, named CFC networks, to separately predict negative expressions and the other kinds of expressions. 
Savchenko \textit{et al.}~\cite{EXPR4} develop a real-time framework through a lightweight EfficientNet model, which may be even implemented for video analytics on mobile devices. 
Phan \textit{et al.}~\cite{EXPR5} employs the pre-trained model RegNet \cite{radosavovic2020designing} as the backbone and then introduces Multi-Head Attention and Transformer Encoder to generate sequence representations.
Kim \textit{et al.} \cite{EXPR6} develop a three-stream network consisting of a visual stream, a temporal stream, and an audio stream, based on Swin transformer \cite{liu2021swin} as the backbone.



\section{Method}
The architecture of the proposed transformer-based multimodal framework is shown in ~\cref{fig:pipeline}. The whole framework consists of a static vision feature extractor, three dynamic multimodal feature extractors and a transformer-based fusion module. 
The static vision feature extractor captures the static vision expression features from the current frame.
Also, the sequences of the cropped face frames, audio, and spoken words from the two seconds around the current frame are sent into the corresponding modal feature extractor to provide more dynamic emotion information that facilitates AU detection and expression recognition. 
Then, the proposed transformer-based fusion module is responsible for integrating the static vision feature and the dynamic multimodal information.
The output of each modal fusion module and the static vision feature are concatenated for the final recognition task. Besides, to alleviate the imbalanced problem, we use the resample strategy and remix~\cite{chou2020remix} data augmentation in the training process. A filter strategy is also utilized for smoothing the predictions.
\begin{figure}
    \centering
    \includegraphics[ width=1\linewidth]{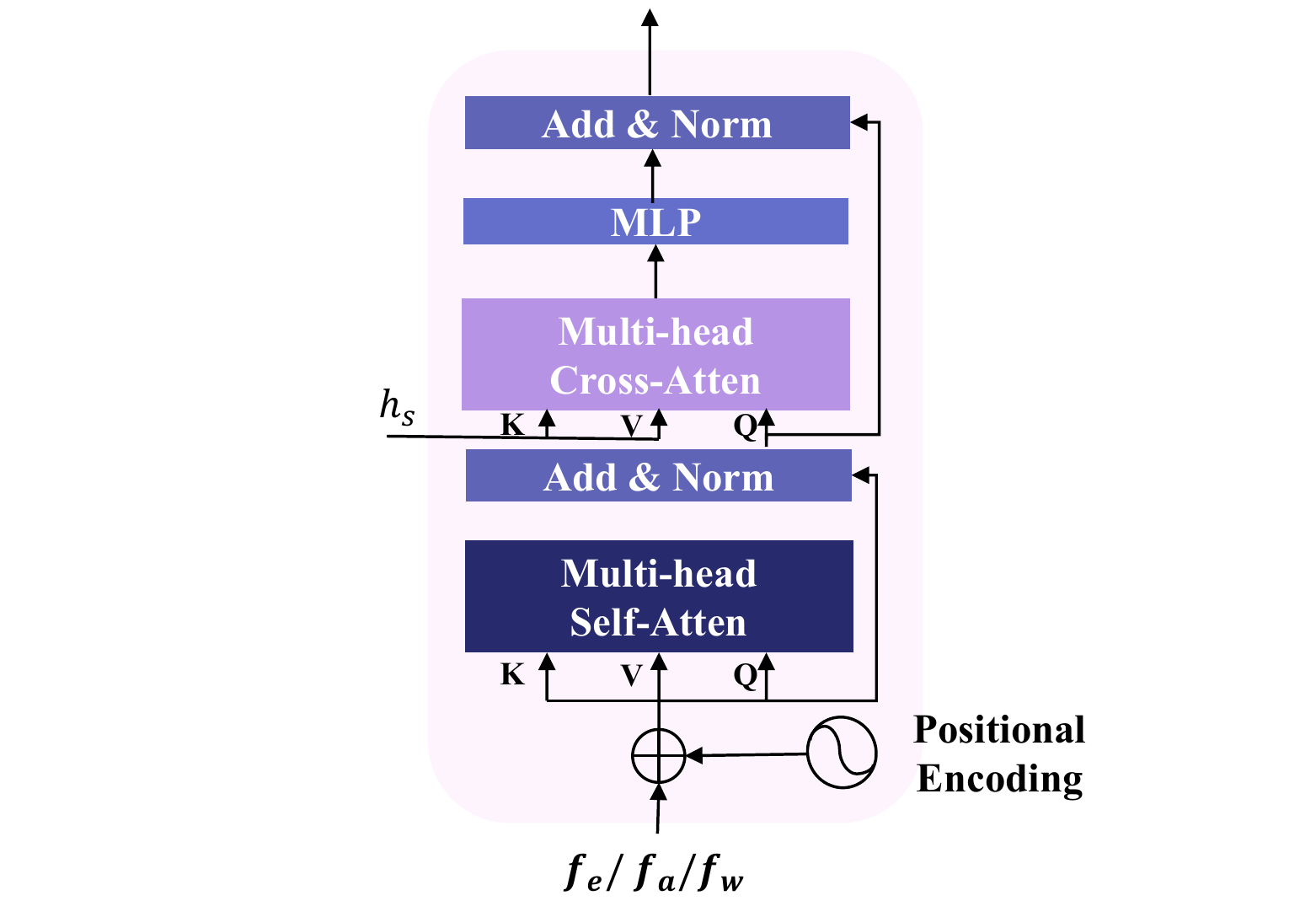}
    \caption{The structure of the transformer-based fusion module. Different from the original transformer decoder, its second multi-head cross attention uses the static vision feature $h_s$ as $K$ and $V$. }
    \label{fig:transformer}
\end{figure}


\subsection{Multimodal feature extraction}
It is commonly acknowledged that human usually conveys and perceives emotion state from vision, audio and text. Based on this observation, it is a straightforward idea to combine multimodal features for full emotion information collection. 
The proposed framework takes advantage of the three modal information and extracts the corresponding original features from some pretrained models or open-sourced tools.

Specifically, the static vision feature extractor with the structure of Inception-Resnet~\cite{szegedy2017inception} is used to capture the vision features $h_s$ of the current frame for its affective analysis. 
To obtain an effective parameter initialization, we make our static vision feature extractor pretrained on the expression embedding task~\cite{zhang2021learning} by the distillation. It is proven that the expression embedding can represent the complex and fine-grained human expressions~\cite{zhang2021learning}. It is of great value to expression-related tasks like AU detection and expression recognition.

Besides, we consider the dynamic multimodal information. We choose $M$ adjacent frames before and after the current frame and compute their expression embedding features by the pretrained DLN network~\cite{zhang2021learning}. In our experiments, we set $M$ to 60. The output expression embedding sequence is represented as $r_{e}$.
Likely, we extract the MFCC, pitch, and short-time energy features from the audio during these frames and concatenate them as audio features sequence $r_a$. 
We choose these kinds of raw audio features due to their robustness.
Also, the spoken words during these frames can be obtained by the open-source ASR tools~\cite{speechbrain} and they are encoded by pretrained Bert model~\cite{devlin2018bert}. $r_w$ is the output sequence of the word embedding extractor.

\subsection{Transformer-based fusion module}
To effectively integrate the static and dynamic multimodal features, we propose a transformer-based fusion module that makes different modal sequence features focus on the current frame detection task. 
More concretely, $r_{e}$, $r_a$ and $r_w$ from the dynamic multimodal feature extractors are first encoded by 4 GRU~\cite{chung2014empirical} for capturing their individual dynamic features $h_{e}$, $h_a$ and $h_w$.

Then, we choose the decoder structure of transformer~\cite{vaswani2017attention} as our fusion module~(See~\cref{fig:transformer}) to integrate the static vision feature $h_s$ and other dynamic modal features. As shown in \cref{fig:transformer}, different from the original transformer decoder, the $K$ and $V$ of the second multi-head attention are replaced by the $h_s$. As a result, the output of the second multi-head attention can be followed:

\begin{equation}
 \begin{split}
    Attention(Q,K,V)  &= Attention(h^{'}_{(.)},h_s,h_s) \\
                      &= softmax(\frac{h^{'}_{(.)}h_s^{T}}{\sqrt{d_k} })h_s 
 \end{split}
\end{equation}
Where $d_k$ is the dimension of $h_s$, $h^{'}_{(.)}$ is each modal input feature of the second multi-head cross attention module. The correlation between the current frame static vision information and the uni-modal dynamic feature can be built by the attention computation. The output integrated features can focus on the key point parts for the current frame detection task.


\subsection{Training}
Except for the framework structure, some network details and training tricks are introduced in this part.

\noindent \textbf{Resampling}.
Due to the serious imbalance of the expression recognition training set, we resample the training samples from the original dataset. 
We observe that both category imbalance and identity imbalance exist in the training dataset. 
If a video contains more than N frames of a particular class, we randomly sample N frames of that class from the video. We set N to 200 and 50 for the minority classes (Anger, Disgust, Fear, Sadness, and Surprise ) and the majority classes (Neutral, Happiness, and Other) respectively. 
In this way, we can avoid the model overfitting to the specific identities and the majority classes.

\noindent \textbf{Data augmentation}. Except for the commonly-used data augmentation operations like rotation, color jitter, center crop, Flip, etc, we take advantages of remix~\cite{chou2020remix} that considers the imbalanced categories based on the original mixup~\cite{zhang2017mixup}. It controls the soft labels of mixed samples more biased towards the minority classes. The process can be followed:
\begin{equation}
 \begin{split}
    x_{new} = \lambda x_i + (1-\lambda) x_j \\
    y_{new} = \lambda_y y_i + (1-\lambda_y) y_j
 \end{split}
\end{equation}
In particular, the input of the model~($x_{new}$) is the mixed sample that is linearly combined from an arbitrary sample pair~($x_i$ and $x_j$) by a mixing factor $\lambda$. $\lambda$ is sampled from the beta distribution. Different from mixup~\cite{zhang2017mixup}, $\lambda_y$ is related to the labels of the sample pair. The rule for $\lambda_y$ is:
\begin{equation}
 \begin{split}
    \lambda_y = \left\{\begin{matrix}
 0  &  n_i/n_j\ge \kappa \ and \ \lambda < \tau \\
 1  &  n_i/n_j\le 1/\kappa \ and \ 1-\lambda < \tau\\
 \lambda & otherwise
\end{matrix}\right.
 \end{split}
\end{equation}
where $n_i$ and $n_j$ is the number of samples with the label of $y_i$ and $y_j$. $\kappa$ and $\tau$ are two hyper-parameters and are usually set as $3$ and $0.5$. In this way, when one mixed sample is from one minority class and one majority class, the mixed label will be dominated by the minority class. 

\noindent \textbf{Post-process}. It can be observed that expression labels seldom rapidly change in a continuous sequence of frames. Therefore, we use the filter operation to smooth the model prediction results, removing some glitches in the temporal sequences. 
In particular, for each hopping point in the sequences of predictions, we first calculate the number of consecutive frames with the same prediction labels. If the number of consecutive frames is less than the threshold, we consider these points as outliers and count the most common values in adjacent frames to correct the hopping points. In addition, due to the data imbalance, we use a lower threshold and smaller adjacent window length for the minority classes in Aff-Wild2~\cite{kollias2022abaw} dataset.

\noindent \textbf{Loss function}. For AU detection, we utilize the cross-entropy loss function $\mathcal{L}_{\textit{AU\_CE}}$ and circle loss function $\mathcal{L}_{\textit{AU\_Circle}}$ like the work~\cite{zhang2021prior} for training. For expression recognition, we use the cross-entropy loss $\mathcal{L}_{\textit{EXPR\_CE}}$ for training. 
In AU detection task, the direct output of the MLP is $\mathcal{S} = \{{s}_{1}, {s}_{2},..., {s}_{12}\}\in\mathbb{R}^{12}$ without scaling. The AU probability~$\mathcal{\hat{Y}}= \{\hat{y}_{1}, \hat{y}_{2},...,\hat{y}_{12}\}$ can be computed by sigmoid activation function for the output $\mathcal{S}$, and the ground-truth binary AU label is $\mathcal{Y}=\{y_{1}, y_{2},...,y_{12}\}\in \mathbb{B}^{12}, \mathbb{B}=\{0,1\}$, where $1$ denotes the corresponding action unit is activated and vice versa. The loss functions for AU task can be formulated as: 

\begin{equation}
    \mathcal{L}_{\textit{AU\_CE}}=-\frac{1}{12}\sum_{j=1}^{12}[y_{j}\log\hat{y}_{j} 
     + (1-y_{j})\log(1-\hat{y}_{j})].
\end{equation}

\begin{equation}
 \begin{split}
    \mathcal{L}_{\textit{AU\_Circle}}&=\log(1+\sum_{i\in \Omega_{0}}e^{s_{i}})+\log(1+\sum_{j\in \Omega_{1}}e^{-s_{j}}),\\
     \Omega_0 &= \{~i~~ |~~ \text{if}~~ y_i=0~\},\\
     \Omega_1 &= \{~j~~ |~~ \text{if}~~ y_j=1~\}.
 \end{split}
\end{equation} 

In EXPR task, the softmax output of the MLP is $\mathcal{\hat{Z}}= \{\hat{z}_{1}, \hat{z}_{2},...,\hat{z}_{8}\}$, and the one-hot ground-truth EXPR label is $\mathcal{Z}=\{z_{1}, z_{2},...,z_{8}\}\in \mathbb{C}^{8}, \mathbb{C}=\{0,1\}$. The loss function for EXPR task can be formulated as: 

\begin{equation}
    \mathcal{L}_{\textit{EXPR\_CE}}=-\frac{1}{8}\sum_{j=1}^{8}z_{j}\log\hat{z}_{j}.
\end{equation}


\begin{table*}[!t]

\renewcommand\arraystretch{1.05} 
\centering
{

  \begin{tabular}{l|cccccccccccc|c}
    \hline
    \textbf{Val Set} &\textbf{AU1} &\textbf{AU2} &\textbf{AU4} &\textbf{AU6} &\textbf{AU7} &\textbf{AU10} &\textbf{AU12} &\textbf{AU15} &\textbf{AU23} &\textbf{AU24} &\textbf{AU25} &\textbf{AU26} &\textbf{Avg.}\\[1pt]
    \hline \hline
    Official &55.3 & \textbf{48.9} & 56.7 & \textbf{62.8} & 74.4 & 75.5 & 73.6 & 28.1 & 10.5 & 20.8 & 83.9 & 39.1 & 52.5\\
    fold-1   &60.2 & 55.1 & 58.6 & 65.3 & 74.8 & \textbf{77.8} & 71.8 & 24.2 & \textbf{13.8} & 26.3 & \textbf{87.7} & \textbf{39.6} & \textbf{54.6}\\
    fold-2   &\textbf{66.9} & 54.3 & \textbf{66.1} & 62.8 & 78.9 & \textbf{71.8} & 73.6 & \textbf{15.2} & \textbf{7.80} & \textbf{31.6} & 85.4 & 38.4 & 54.4\\
    fold-3   &59.1 & 51.9 & 59.3 & 67.8 & \textbf{79.3} & 73.6 & \textbf{74.5} & 27.1 & 12.9 & \textbf{16.2} & 84.9 & 35.4 & 53.5 \\
    fold-4   &\textbf{54.7} & 49.9 & 55.2 & \textbf{69.6} & \textbf{73.1} & 74.4 & 72.9 & \textbf{30.2} & 10.8 & 21.5 & \textbf{82.0} & \textbf{30.9} & \textbf{52.1} \\
    fold-5   &62.3 & \textbf{57.4} & \textbf{54.2} & 64.5 & 76.2 & 76.1 & \textbf{72.3} & 28.7 & 11.8 & 19.4 & 86.8 & 33.7 & 53.6 \\
    
    \hline
  \end{tabular}}
  \caption{The AU F1 scores~(in \%) of models that are trained and tested on different folds (including the original training/validation set of \textit{Aff-Wild2} dataset). The highest and lowest scores are both indicated in bold.}
  \label{tab:AU_F1_val}
\end{table*}

\begin{table*}[!t]

\renewcommand\arraystretch{1.05} 
\centering
{

  \begin{tabular}{l|cccccccc|c}
    \hline
    \textbf{Val Set} &\textbf{Neutral} &\textbf{Anger} &\textbf{Disgust} &\textbf{Fear} &\textbf{Happiness} &\textbf{Sadness} &\textbf{Surprise} &\textbf{Other} &\textbf{Avg.}\\[1pt]
    \hline \hline
    Official&  57.6 & 34.7 & 14.0 & 20.0 & 53.3 & \textbf{43.0} & 37.9 & \textbf{54.7} & 39.4 \\
    fold-1 &   54.6 & \textbf{39.1} & 10.0 & 23.3 & \textbf{50.6} & 34.7 & 35.3 & 55.9 & 37.9 \\
    fold-2 &   61.4 & \textbf{8.10} & \textbf{31.7} & \textbf{38.7} & 60.2 & 33.4 & 40.2 & 55.1 & \textbf{41.1} \\
    fold-3  &  \textbf{65.8} & 29.9 & 19.8 & 22.0 & 51.7 & 23.6 & \textbf{27.8} & \textbf{61.9} & 37.8 \\
    fold-4 &   \textbf{54.2} & 33.9 & 6.0 & 14.1 & \textbf{60.3} & \textbf{21.9} & \textbf{48.5} & 59.6 & 37.3 \\
    fold-5 &   56.4 & 25.7 & \textbf{5.0} & \textbf{8.1} & 60.0 & 29.7 & 41.8 & 62.3 & \textbf{36.1} \\
    
    \hline
  \end{tabular}}
  \caption{The expression F1 scores~(in \%) of models that are trained and tested on different folds (including the original training/validation set of \textit{Aff-Wild2} dataset). The highest and lowest scores are both indicated in bold.}
  \label{tab:exp_F1_val}
\end{table*}
\section{Experiments}
In this section, we will first introduce our used datasets and the implementation details. 
Then we evaluate our model on the ABAW3 competition metrics. 
To prove the effectiveness of each module in our framework, We also present several ablation studies. 
\subsection{Datasets}
The 3rd Workshop and Competition on Affective Behavior Analysis in-the-wild provides the Aff-wild2 datasets as the official datasets. It contains 564 in-the-wild videos and around 2.8M frames from Youtube. Most of them show human spontaneous affective behaviors. 
In detail, there are 548 videos in Aff-wild2~\cite{kollias2022abaw} annotated with 12 AUs, namely AU1, AU2, AU4, AU6, AU7, AU10, AU12, AU15, AU23, AU24, AU25, and AU26. The ABAW3 competition provides 295 of them as the trainset, 105 of them as the validation set, and 141 of them as the test set. 
Aff-wild2~\cite{kollias2022abaw} contains 548 videos annotated with discrete expression categories, namely neutral, happiness, anger, disgust, fear, sadness and surprise, and other. Different from the ABAW2~\cite{kollias2021analysing}, the new added other class means the expression state is different from the other seven basic expressions. The ABAW3 competition provides 248 of them as the trainset, 70 of them as the validation set, and 228 of them as the test set. 
Aff-wild2~\cite{kollias2022abaw} contains 567 videos annotated with valence and arousal. The ABAW3 competition provides 341 of them as the trainset, 71 of them as the validation set, and 152 of them as the test set. 
ABAW3 creates a static version of Aff-wild2~\cite{kollias2022abaw} s-Aff-wild2 that is used in the Multi-task learning challenge. The ABAW3 competition provides 145,273 frames as the trainset, 27,087 frames as the validation set, and 51,245 frames as the test set. 
The official use F1 score as the metric for AU detection and expression recognition and Concordance Correlation Coefficient as the metric for VA detection. 
We use the official trainset for training and evaluate our model on the official validation set in the AU and EXPR tracks. Final scores of ABAW3 competition are computed on the unseen official test set.

\subsection{Implementation details}
We use OpenCV to process all videos in the \textit{Aff-Wild2} dataset~\cite{kollias2019expression} into frames and crop all facial images into $224\times 224$ scale by OpenFace~\cite{baltrusaitis2018openface} detector. 
The static vision feature extractor is initialized by the pretrained expression embedding model distilled from DLN~\cite{zhang2021learning}.
The audio feature extraction relies on the open-sourced tool librosa~\cite{mcfee2015librosa}.
We obtain the spoken words by the open-sourced ASR tool~\cite{speechbrain} and extract their word embedding by the pretrained Bert~\cite{devlin2018bert} model.
Our training process is implemented based on PyTorch. The training procedure runs about 20 epochs on an NVIDIA RTX 3090 graphics card with a learning rate of 0.002 and batch size of 80. We use a stochastic gradient (SGD) optimizer with a cosine annealing warm restart learning rate scheduler.

\begin{table*}[!t]
\renewcommand\arraystretch{1.05} 
\centering
{
  
  \begin{tabular}{l|cccccccccccc|c}
    \hline
    \textbf{Condition} &\textbf{AU1} &\textbf{AU2} &\textbf{AU4} &\textbf{AU6} &\textbf{AU7} &\textbf{AU10} &\textbf{AU12} &\textbf{AU15} &\textbf{AU23} &\textbf{AU24} &\textbf{AU25} &\textbf{AU26} &\textbf{Avg.}\\[1pt]
    \hline \hline
    only static & 54.0 & 44.9 & 53.1 & 62.9 & 73.1 & 74.1 & 74.3 & 26.7 & 13.0 & 16.8 & 82.2 & 35.5 & 50.9 \\
    w/o exp\_emb & 53.3 & 42.8 & 55.3 & 62.3 & 75.0 & 75.2 & \textbf{74.8} & 28.2 & 11.1 & 19.8 & 83.8 & 32.7 & 51.2 \\
    w/o audio  & 55.2 & 49.6 & \textbf{57.4} & \textbf{63.7} & 75.1 & 75.1 & 74.1 & 15.0 & 4.20 & 21.6 & 83.7 & \textbf{40.0} & 51.2\\
    w/o word   & \textbf{56.9} & \textbf{49.7} & 55.0 & 62.8 & \textbf{75.5} & 72.3 & 73.1 & 15.3 & 7.70 & \textbf{28.6} & 74.6 & 37.9 & 50.8\\
    w/o trans  & 53.5 & 42.2 & 54.6 & 62.0 & 73.8 & 74.0 & 73.3 & \textbf{29.9} & \textbf{16.5} & 24.0 & 82.9 & 33.1 & 51.7 \\
    Ours & 55.3 & 48.9 & 56.7 & 62.8 & 74.4 & \textbf{75.5} & 73.6 & 28.1 & 10.5 & 20.8 & \textbf{83.9} & 39.1 & \textbf{52.5}\\
    
    \hline
  \end{tabular}}
  \caption{Ablation study results of different modality features and transformer-based fusion module for AU detection task. All scores are computed based on the official validation set.The highest is indicated in bold.}
  \label{tab:AU_abla}
\end{table*}

\begin{table*}[!t]
\renewcommand\arraystretch{1.05} 
\centering
{
  
  \begin{tabular}{l|cccccccc|c}
    \hline
    \textbf{Condition} &\textbf{Neutral} &\textbf{Anger} &\textbf{Disgust} &\textbf{Fear} &\textbf{Happiness} &\textbf{Sadness} &\textbf{Surprise} &\textbf{Other} &\textbf{Avg.}\\[1pt]
    \hline \hline
    only static & 62.1 & 27.7 & 2.32 & 3.98 & 49.1 & 15.3 & 30.1 & 45.4 & 29.5 \\ 
    w/o exp\_emb & 62.7 & \textbf{45.6} & 10.3 & 11.5 & 51.7 & 17.9 & 30.3 & \textbf{57.4} & 35.9 \\
    w/o audio   & \textbf{65.3} & 41.9 & 5.06 & 10.7 & 54.4 & 32.0 & 30.2 & 50.6 & 36.3 \\
    w/o word     & 60.8 & 45.1 & 2.20 & 18.8 & 48.3 & 23.3 & 25.8 & 36.4 & 32.6 \\
    w/o trans   & 64.2 & 33.6 & 3.00 & \textbf{23.3} &  \textbf{55.0} & 28.4 & 31.0 & 52.5 & 36.4 \\
    Ours         & 57.6 & 34.7 & \textbf{14.0} & 20.0 & 53.3 & \textbf{43.0} & \textbf{37.9} & 54.7 & \textbf{39.4} \\
    
    \hline
  \end{tabular}}
  \caption{Ablation study results of different modality features and transformer-based fusion module for Expression recognition task. All scores are computed based on the official validation set. The highest is indicated in bold.}
  \label{tab:EXP_abla}
\end{table*}
\begin{table}[!t]
\renewcommand\arraystretch{1.05} 
\centering
{

  \begin{tabular}{l|c}
    \hline
    \textbf{Method} &\textbf{AU Avg F1 score} \\[1pt]
    \hline \hline
    Tallec \textit{et al.}\cite{AU_6} & 44.3 \\
    Savchenko \textit{et al.}\cite{EXPR4} & 47.3 \\
    Wang \textit{et al.}\cite{AU_4} & 48.8 \\
    Nguyen \textit{et al.}\cite{AU_3} & 49.0 \\
    Jiang \textit{et al.}\cite{AU_2} & 49.8 \\
    our method & \textbf{49.9} \\
    \hline
  \end{tabular}}
  \caption{The AU F1 scores~(in \%) of different models on the official Aff-wild2 test set. The highest is indicated in bold.}
  \label{tab:AU_com}
\end{table}

\begin{table}[!t]
\renewcommand\arraystretch{1.05} 
\centering
{

  \begin{tabular}{l|c}
    \hline
    \textbf{Method} &\textbf{EXPR Avg F1 score} \\[1pt]
    \hline \hline
    Kim \textit{et al.}\cite{EXPR6} & 27.2 \\
    Phan \textit{et al.}\cite{EXPR5} & 28.6 \\
    Savchenko \textit{et al.}\cite{EXPR4} & 30.3 \\
    Xue \textit{et al.} \cite{EXPR3} & 32.2 \\
    Jeong \textit{et al.}\cite{EXPR2} & 33.8 \\
    our method & \textbf{35.9} \\
    \hline
  \end{tabular}}
  \caption{The expression F1 scores~(in \%) of different models on the official Aff-wild2 test set. The highest is indicated in bold.}
  \label{tab:exp_com}
\end{table}
\subsection{Results on the validation set}
The official ABAW3 provides the official training/validation/test set based on the videos. To relieve the disturbance caused by different dataset divisions, we also perform the 5-fold random cross-validation. \cref{tab:AU_F1_val} and \cref{tab:exp_F1_val} presents the AU detection results of the official validation set and our 5-fold validation sets. From the above table, it can be observed that the different dataset division results in the fluctuating average F1 score, especially for the expression recognition task. The AU average F1 score ranges from $52.1\%$ to $54.6\%$ while the expression average F1 score ranges from $36.1\%$ to $41.1\%$. 
The reason is that the trainset contains limited identities, especially for the minority classes. To alleviate the bad influence, we make the final decisions for the test set prediction by voting on these models using different dataset divisions.

\subsection{Results on the test set}
We present the final results on the official test set for AU detection and expression recognition task in the ~\cref{tab:AU_com} and ~\cref{tab:exp_com}. In particular, our method achieves $49.9\%$ F1 score in AU track and $35.9\%$ F1 score in EXPR track, winning the first prizes in the AU and EXPR tracks.

Observing these methods, Transformer structure is usually used. 
Tallec \textit{et al.}~\cite{AU_6} and Nguyen \textit{et al.}~\cite{AU_3} use it on the vision feature for AU detection. Phan \textit{et al.}~\cite{EXPR5} use it for expression recognition. 
Wang \textit{et al.}~\cite{AU_4} use the transformer encoder structure to model the relationships between different AUs. 
Also, it is practical to use extra datasets for pretraining~\cite{AU_2,EXPR4}. 
Xue \textit{et al.}~\cite{EXPR3} propose a Coarse-to-Fine network to improve the prediction for the hard negative expression classes.

Different from them, we use more valid multimodal features. Also, our transformer decoder structure is for multimodal feature fusion.

\subsection{Ablation study}
To prove the effectiveness of the multimodal features and our transformer-based fusion module, we conduct ablation studies by comparing the models trained without the corresponding features or modules. We present the quantitative results for AU detection and expression recognition in \cref{tab:AU_abla} and \cref{tab:EXP_abla}. As shown in the tables, each modality and our fusion module can facilitate AU detection and expression recognition.

\noindent\textbf{Multimodal features.} To indicate the effectiveness of the dynamic multimodal features, we compare our complete model with the model that only uses static vision features~(\textit{only static}). From the \cref{tab:AU_abla} and \cref{tab:EXP_abla}, it can be observed that multimodal features bring about a distinctive improvement for both AU and expression tasks. Next, we will give a detailed analysis of each modality's features. 

To verify the benefits of the expression embedding features, we conduct an ablation study by removing the expression embedding extractor~(\textit{w/o exp\_emb}). From the \cref{tab:AU_abla} and \cref{tab:EXP_abla}, it can be observed that the expression embedding feature plays an important role in AU and expression tasks. Without it, AU average F1 score falls from 52.5\% to 51.2\% and the expression average F1 score falls from 39.4\% to 35.9\%.

To prove the effectiveness of the word embedding features, we compare our complete model with the model without the word embedding extractor~(\textit{w/o word}). As shown in \cref{tab:AU_abla} and \cref{tab:EXP_abla}, without word feature, AU average F1 score falls from 52.5\% to 50.8\% and expression average F1 score falls from 39.4\% to 32.6\%. It can be observed that word embedding features play a more important role in expression recognition than AU detection.

Likely, we conduct the ablation study by removing the audio features extractor~(\textit{w/o audio}) to prove the effectiveness of the MFCC, pitch, and energy features. As shown in \cref{tab:AU_abla} and \cref{tab:EXP_abla}, without audio features, AU average F1 score falls from 52.5\% to 51.2\% and expression average F1 score falls from 39.4\% to 36.3\%. It can be indicated that audio information can facilitate the AU and expression tasks.

\noindent \textbf{Transformer-based fusion module.} We perform an ablation study by replacing the transformer-based fusion module with concatenation to evaluate its significance. In the \cref{tab:AU_abla} and \cref{tab:EXP_abla}, without the fusion module, AU average F1 score falls from 52.5\% to 51.7\% and expression average F1 score falls from 39.4\% to 36.4\%. This proves that the transformer-based fusion module promotes the model performance effectively. In addition, the rising degree of F1 score brought by the fusion module is slightly less than multimodal features overall. This indicates that multimodal features are more important.


\section{Conclusion}
In this paper, we introduce our transformer-based multimodal information fusion framework for AU detection and expression recognition in the ABAW3 Competition. We propose to exploit the static vision feature and three kinds of dynamic multimodal features to fully collect the human emotion clues. Besides, to integrate the static features and dynamic multimodal features, we utilize the transformer decoder structure. In participating in the ABAW3 competition, we won the first prizes in the AU track and EXPR track. The competition results prove the superiority of our method. Also, the quantities ablation studies indicate that each multimodal feature and fusion module can improve the model performance for affective tasks.

{\small
\bibliographystyle{ieee_fullname}
\bibliography{egbib}
}

\end{document}